\documentclass[sigconf]{acmart}
\usepackage{graphicx}
\usepackage{multirow}
\usepackage{amsmath}
\usepackage{amsfonts}
\usepackage{subfig}
\usepackage{booktabs}
\usepackage{multicol}
\usepackage{gensymb}
\usepackage{multirow}

\usepackage[ruled,vlined,norelsize,\languagename]{algorithm2e}
\setcopyright{rightsretained} 
\renewcommand\footnotetextcopyrightpermission[1]{} 
\begin{document}

\title{Combining Off and On-Policy Training in Model-Based Reinforcement Learning}

\author{Alexandre Borges}
\affiliation{%
  \institution{INESC-ID / Instituto Superior Técnico \\University of Lisbon}
  \city{Lisbon}
  \country{Portugal}
}
\author{Arlindo L. Oliveira}
\affiliation{%
  \institution{INESC-ID / Instituto Superior Técnico \\University of Lisbon}
  \city{Lisbon}
  \country{Portugal}
}

\renewcommand{\shortauthors}{A. Borges}%

\begin{abstract}
The combination of deep learning and Monte Carlo Tree Search (MCTS) has shown to be effective in various domains, such as board and video games. AlphaGo \cite{Silver2016} represented a significant step forward in our ability to learn complex board games, and it was rapidly followed by significant advances, such as AlphaGo Zero \cite{Silver2017} and AlphaZero \cite{Silver2018}. Recently,  MuZero \cite{Schrittwieser2020} demonstrated that it is possible to master both Atari games and board games by directly learning a model of the environment, which is then used with Monte Carlo Tree Search (MCTS) \cite{Coulom2007}  to decide what move to play in each position. During tree search, the algorithm simulates games by exploring several possible moves and then picks the action that corresponds to the most promising trajectory. When training, limited use is made of these simulated games since none of their trajectories are directly used as training examples. Even if we consider that not all trajectories from simulated games are useful, there are thousands of potentially useful trajectories that are discarded. Using information from these trajectories would provide more training data, more quickly, leading to faster convergence and higher sample efficiency. Recent work \cite{Willemsen2020} introduced an off-policy value target for AlphaZero that uses data from simulated games. In this work, we propose a way to obtain off-policy targets using data from simulated games in MuZero. We combine these off-policy targets with the on-policy targets already used in MuZero in several ways, and study the impact of these targets and their combinations in three environments with distinct characteristics. When used in the right combinations, our results show that these targets can speed up the training process and lead to faster convergence and higher rewards than the ones obtained by MuZero. 
\end{abstract}

\keywords{Deep Reinforcement Learning, Model-Based Learning, MCTS, MuZero, Off-Policy Learning, On-Policy Learning}

\maketitle

\section{Introduction}
Board games have often been solved using planning algorithms, more specifically, using tree search with handcrafted heuristics. TD-Gammon \cite{Tesauro1995} demonstrated that it is possible to learn a position evaluation function and a policy from self-play to guide the tree search instead of using handcrafted heuristics and achieved super-human performance in the game of backgammon. However, the success of TD-Gammon was hard to replicate in more complex games such as chess. When Deep Blue \cite{Campbell2002} was able to beat the then world chess champion Garry Kasparov, it still used tree search guided by handcrafted heuristics. These approaches failed for even more complex games, such as Go, because of its high branching factor and game length. AlphaGo \cite{Silver2016} was the first computer program to beat a professional human player in the game of Go by combining deep neural networks with tree search. In the first training phase, the system learned from expert games, but a second training phase enabled the system to improve its performance by self-play using reinforcement learning. AlphaGoZero \cite{Silver2017} learned only through self-play and was able to beat AlphaGo soundly.
AlphaZero \cite{Silver2018} improved on AlphaGoZero by generalizing the model to any board game. However, these methods were designed only for board games, specifically for two-player zero-sum games.

Recently, MuZero \cite{Schrittwieser2020} improved on AlphaZero by generalizing it even more, so that it can learn to play singe-agent games while, at the same time, learning a model of the environment. MuZero is more flexible than any of its predecessors and can both master Atari games and board games. It can also be used in environments where we do not have access to a simulator of the environment. 

\textbf{Motivation.}
All these algorithms, regardless of their successes, are costly to train. AlphaZero took three days to achieve super-human performance in the game of Go, using a total of 5064 TPUs (Tensor Processing Units). MuZero, while being more efficient, still used 40 TPU for Atari Games and 1016 TPU for board games. 

Both AlphaZero and MuZero use MCTS \cite{Coulom2007} to decide what move to play in each position. During tree search, the algorithm simulates games by exploring several possible moves and then selects the action that corresponds to the most promising trajectory. Even though not all trajectories from these simulated games correspond to good moves, they contain information useful for training. Therefore, these trajectories could provide more data to the learning system, enabling it to learn more quickly and leading to faster convergence and higher sample efficiency. 

In this work, inspired by recent work \cite{Willemsen2020} that introduced an off-policy value target for AlphaZero, we present three main contributions:
\begin{itemize}
    \item We propose a way to obtain off-policy targets by using data from the MCTS tree in MuZero.
    \item We combine these off-policy targets with the on-policy targets already used in MuZero in several ways.
    \item We study the impact of using these targets and their combinations in three environments with distinct characteristics.
\end{itemize}

The rest of this work is organized as follows. Section \ref{sec:bg} presents the background and related work of relevance for the proposed extensions. Section \ref{sec:mz-gb} explains the proposed extensions and Section \ref{sec:eval} presents the results. Finally, Section \ref{sec:con} presents the main takeaway messages and points to possible directions for future work.

\section{Background and Related Work}
\label{sec:bg}
\subsection{Reinforcement Learning}
Reinforcement learning is an area of machine learning that deals with the task of sequential decision making. In these problems, we have an agent that learns through interactions in an environment. We can describe the environment as a Markov Decision Process (MDP). An MDP is a 5-tuple  (\(\mathcal{S}\),  \(\mathcal{A}\), \(P\), \(R\), \(\gamma\)) where: \(\mathcal{S}\) is a set of states; \(\mathcal{A}\) is a set of actions; \(P: \mathcal{S} \times \mathcal{A} \times \mathcal{S} \to [0,1]\) is the transition function that determines the probability of transitioning to a state given an action; \(R: \mathcal{S} \times \mathcal{A} \times \mathcal{S} \to  \mathbb{R}\) is the reward function that for a given state \(s_t\), action \(a_t\) and next state \(s_{t+1}\) triplet returns the reward; and \(\gamma \in [0,1)\) is a discounting factor for future rewards.

Interactions with the environment can be broken into episodes. An episode is composed of several timesteps. In each timestep \(t\), the agent observes a state \(s_t\), takes an action \(a_t\) and receives a reward \(r_t\) from the environment which now transitions into a new state \(s_{t+1}\).

The learning objective for the agent is to maximize the reward over the long run. The agent can solve a reinforcement learning problem by learning one or more of the following things: a policy \(\pi(s)\); a value function \(V(s)\); or a model of the environment. One or more of these can then be used to plan a course of action that maximizes the reward. A policy \(\pi: \mathcal{S} \times \mathcal{A} \to [0,1] \) determines the probability that the agent will take an action in a particular state. A value function \(V_\pi: \mathcal{S} \to \mathbb{R}\) evaluates how good a state is based on the expected discounted sum of the rewards for a certain policy, \( \mathbb{E}_\pi [\sum_{k=0}^{\infty} y^k r_{t+k+1} | S_t = s]\). A model of the environment includes the transition function \(P\) between states and the reward function \(R\) for each state.

Another function of interest is the action-value function \(Q_\pi: \mathcal{S} \times \mathcal{A}  \to \mathbb{R} \) that represents the value of taking action \(a\) in state \(s\) and then following policy \(\pi\) for the remaining timesteps.

Model-free algorithms are algorithms that are value and/or policy-based, meaning that they learn a value function and/or a policy. Algorithms that are dependent on a model of the environment are said to be model-based.

The agent interacts with the environment using a certain policy called the behaviour policy. When training, the model can update its estimates by either using the behavior policy or some other policy. On-policy learning occurs when the behaviour policy and the training policy are the same, and off-policy learning occurs when they are different.

\subsection{Combining On-policy and Off-policy Learning }
Most of the work that combines off and on-policy learning combines policy-based methods with value-based methods, for instance combining on-policy policy gradient with off-policy value functions \cite{Gu2017} \cite{Gu2016} \cite{Lehnert2015} \cite{Hu2019}. 


Backward Q-learning \cite{Wang2013} combines SARSA \cite{Rummery1994} updates with Q-learning \cite{Watkins1992} updates. When collecting episodes, the Q-value estimate is updated using the SARSA update. Afterwards, when a terminal state is reached, the trajectory is followed backwards and a Q-learning update is performed.
 
Backward Q-learning uses tabular methods, explicitly storing the relevant values in tables. When it comes to deep reinforcement learning, Q-learning targets have been combined with Monte Carlo targets \cite{Hausknecht2016}:
\begin{equation}
  y = \beta y_{on\_policy\_MC} + (1 - \beta)y_{q\_learning} 
\end{equation}
 where \(y_{on\_policy\_MC}\) is calculated directly from the rewards of complete episodes in the replay buffer, \(y_{q\_learning}\) is a 1-step Q-learning target, and \(\beta\) is a parameter to control mixing between targets. The authors tested it in Deep Q-Learning (DQN) \cite{Mnih2013}  and Deep Deterministic Policy Gradient (DDPG) \cite{Lillicrap2015}. This method improved learning and stability in DDPG. However, in DQN it hindered training across four out of five Atari games. 
 
\subsection{MuZero} \label{sec:mz}
MuZero learns a model of the environment that is then used for planning. The model is trained to predict the most relevant values for planning which in this case are the reward, the policy and the value function.      

\subsubsection{\textbf{Network}}
MuZero uses three functions to be able to model the dynamics of the environment and to plan. 

\begin{itemize}
    \item \textbf{Representation function \(h_\theta\)}: takes as input the past observations  \(o_1, ..., o_t\) and outputs a hidden state \(s^0\) that will be the root node used for planning.
    \item \textbf{Dynamics function \(g_\theta\)}: takes as input the previous hidden state \(s^{k-1}\) and an action \(a^k\), and outputs the next hidden state \(s^k\) and immediate reward \(r^k\) 
    \item \textbf{Prediction function \(f_\theta\)}: this is the same as in AlphaZero. It takes as input a hidden state \(s^k\) and outputs the policy \(p^k\) and value \(v^k\) 
\end{itemize}

Note that there is no requirement to be able to obtain the original observations from the hidden state. The only requirement for the hidden state is that it is represented in such a way as to predict the values necessary for planning accurately.

\subsubsection{\textbf{Training.}}
Data for training can be either generated by self-play in the case of two-player zero sum-games or by interacting with the environment in the case of a general MDP. This data is sent to the replay buffer. A simplified version of how training is done in MuZero is illustrated by Algorithm \ref{alg:training_loop}. 

\label{sec:mz_training}
\begin{algorithm}[ht]
\footnotesize
\DontPrintSemicolon
nn = NN()

replayBuffer = replayBuffer()
\BlankLine
\While{True}{
//Data Generation

\For{$i\gets0$ \KwTo numberOfGames}{
    game = newGame()
    
    gameHistory = gameHistory()
    
    done = False
    \BlankLine
    \While{not done}{
        root = MCTS.simulate(game.observations())

        action = MCTS.selectAction(root.childVisits)

        observation, reward, done = game.step(action)
        
        gameHistory.save(root, observation, reward, action, done)
    }
    \BlankLine
    addTrajectoriesToBuffer(replayBuffer, gameHistory)
    
}
\BlankLine
//Training

\For{$i\gets0$ \KwTo numberOfBatches}{
    trainOneBatch(nn, replayBuffer)
}
}

\caption{Simplified MuZero Training Loop}
\label{alg:training_loop}
\end{algorithm}

Trajectories are sampled from the replay buffer for training. 
A trajectory for a timestep \(t\) consists of the following information: the observation history \(o_1, ..., o_t\), the action history \(a_t, ..., a_{t+K}\), the reward history  \(u_t, ..., u_{t+K} \), the policy history \(\pi_t, ...,\pi_K\) and the value history \(z_t, ..., z_{t+K}\) where \(K\) is the number of unroll steps. 

After sampling a trajectory, the representation function \(h_\theta\) receives the past observations \(o_1, ..., o_t\) and gets the first hidden state \(s_{t}^{0}\). Afterwards, the model is unrolled recurrently for \(K\) steps. To do this, at each step \(k\), the dynamics function \(g_\theta\) receives the previous hidden state \(s_{t}^{k-1}\) and the real action \(a_{t+k}\) and predicts the next hidden state \(s_{t}^{k}\) and the reward \(r_{t}^{k}\). The prediction function \(f_\theta\) then takes the predicted hidden step \(s_{t}^{k}\) as input and predicts the value \(v_{t}^{k}\) and policy \(p_{t}^{k}\).

\textbf{Loss.} After unrolling the model, the parameters of these functions are trained jointly end-to-end to predict the policy, value and reward,
\begin{equation}\label{eq:mz_loss}
\footnotesize
l_t(\theta) = \sum^{k}_{k=0} \Big[l^r(u_{t+k}, r^{k}_{t}) + l^v(z_{t+k}, v^k_{t}) + l^p(\pi_{t+k}, p^k_{t})\Big] + c\parallel\theta\parallel^2,
\end{equation}
where \(u_{t+k}\) is the true reward and \(r^k_t\) is the reward predicted by the dynamics function, \(z_{t+k}\) is either the final reward (if board games) or a n-step return, \(v^k_t\)  is the value predicted by the prediction function,  \(\pi_{t+k} \) is the policy from MCTS, and \(p^k_t\)  is the policy predicted by the prediction function. 

\(l^r\), \(l^v\) and \(l^p\) denote the reward, value and policy loss functions respectively and are defined as follows:
\begin{equation}
\footnotesize
l^r(u, r) = \begin{cases}
    0, & \text{two-player zero-sum games}.\\
    \phi(u)^T log(r), & \text{general MDP},
  \end{cases}
\end{equation}

\begin{equation}
\footnotesize
l^v(z, v) = \begin{cases}
    (z - v)^2, & \text{two-player zero-sum games}.\\
    \phi(z)^T log(v), & \text{general MDP},
  \end{cases}
\end{equation}

\begin{equation}
\footnotesize
l^p(\pi, p) = \pi^T log(p).
\end{equation}

 
Note that when the environment is a general MDP, \(z_{t+k}\) is calculated by bootstrapping n-steps into the future. Therefore, \(z_{t+k} =   u_{t+1} + \gamma u_{t+2} + .. +  \gamma^{n-1}  u_{t+n} + \gamma^{n}v_{t+n}\). 

\subsubsection{\textbf{Monte Carlo Tree Search (MCTS)}} The planning algorithm used by MuZero is MCTS that has been modified to include the network when searching. 

MCTS is a best-first search algorithm. The algorithm starts at a game state \(s\) and is used to obtain a policy \(\pi_{MCTS}(s)\) and a value   \(V_{MCTS}(s)\) for that state. In order to do this, a tree is progressively built by performing several simulations that always start at \(s\). 

Each node in the tree represents a game state \(s\) and each edge a possible action \(a\). Each edge \((s,a)\) in the tree contains  the visit count \(N(s,a)\), the mean action-value \(Q(s,a)\), the probability from the policy \(P(s, a)\), the immediate reward \(R(s,a)\), and the state transition \(S(s, a)\).

\textbf{MCTS Simulations.}
First, we use the representation function \(h_\theta\) to obtain the first hidden state \(s^0\). This hidden state will be the root node from where each simulation will start. In each simulation we will do the following steps: 
\begin{itemize}
    \item \textbf{Selection:} from \(s^0\) until we reach a leaf node \(s^{l}\), we will pick an action \(a^k\) where \(k = 1 ... l\) to traverse the tree based on a modified version of Polynomial Upper Confidence Trees (PUCT) \cite{Auger2013}

\begin{equation}
\footnotesize
     a^k =  argmax_{a'} (Q(s,a') + U(s,a')),
\end{equation}
\begin{equation}
\footnotesize
     U(s, a') =  P(s,a) \frac{\sqrt{ N(s)}}{1 + N(s,a')}(c_1 + log(\frac{\sum_b N(s,b) + c_2 + 1}{c_2})).
\label{eq:mz_u}
\end{equation}
where \(c_1\) and \(c_2\) are used to control the influence of the prior \(P(s, a)\) relative to \(Q(s, a)\)

\item \textbf{Expansion and Evaluation:} When we reach \(s^{l-1}\), after selecting the best action \(a^l\), we use the dynamics function  to obtain the reward and the next state \(s^l\): \(g_{\theta}(s^{l-1},a^l) = s^l, r^l\). Afterwards, we compute the value and policy of that new state \(s^l\) using the prediction function:  \(f_{\theta}(s^{l-1},a^l) = v^l, p^l\).

The values predicted are stored to be used in future simulations.

\item \textbf{Backup:} Since we now consider that the environment can emit intermediate rewards, the value of a state is calculated with an (l - k)-step estimate of the cumulative discounted reward plus the value estimate 

\begin{equation} 
\footnotesize
G^k = (\sum^{l - 1 - k}_{\tau = 0} \gamma^{\tau} r_{k + 1 + \tau}) + \gamma^{l-k}v^{l}.
\label{eq:mz_bu}
\end{equation}

We follow the simulation path backwards and update the statistics for each edge. For k = \(l...1\)
\begin{equation}
\footnotesize
Q(s^{k-1}, a^{k}) = \frac{N(s^{k-1}, a^k) \cdot Q(s^{k-1}, a^k) + G^k}{N(s^{k-1}, a^k) + 1},
\end{equation}

\begin{equation} 
\footnotesize
N(s^{k-1}, a^k) = N(s^{k-1}, a^k) + 1.
\end{equation}

\end{itemize}

The value of \(\pi_{MCTS}(s)\) is then obtained based on the visit count of each action,
\begin{equation}   
\footnotesize
 \pi_{MCTS}(s, a) = \frac{N(s, a)^{1/\tau}}{\sum_b N(s,b)^{1/\tau}}.
 \label{eq:pickmove}
\end{equation}

\section{Off and On-policy Learning in MuZero}
 \label{sec:mz-gb}
We can consider that there are two types of games in MuZero: real games and simulated games. Real games are games where the model interacts with the environment. We use trajectories from these games for training. Simulated games are games that are played when performing MCTS simulations. We use these games to get the best possible action when playing. However, using trajectories from simulated games for training would provide more data which might increase convergence speed and sample efficiency. 

In A0GB \cite{Willemsen2020}, the authors propose a way to use data from simulated games in AlphaZero, with a value target obtained greedily from the MCTS tree. Since in MuZero trajectories are sampled instead of positions, the targets proposed in the literature \cite{Carlsson2019} \cite{Moerland2018} \cite{Willemsen2020} for AlphaZero cannot directly be adapted to MuZero. We can, however, apply the same idea as in A0GB of traversing the tree greedily and using the values from that greedy path as training targets. 

In this section, we define a path and a trajectory as a sequence of actions, observations and rewards. However, a path can have any length while a trajectory has a fixed length which, in the case of MuZero, is the number of unroll steps \(K\). Besides that, there are two axes of timesteps to consider: we have timesteps that occur during the real game denoted with the subscript \(t\), and timesteps that occur during a simulated game denoted with the superscript \(t\). For example, the action \(a_{t}^{2}\) means that we followed the real game until timestep \(t\) and then followed a simulated game until timestep \(2\). Therefore, the total length of this game is \(t+2\).




\subsection{\textbf{Creating Simulated Trajectories}}
\label{sec:creating_sim}
As detailed in Section \ref{sec:mz_training}, a valid MuZero training trajectory has several components. We will now explain how each component is obtained if we were to use a simulated path for an observation at timestep \(t\). 
The observation history \(o_1, ..., o_t\) is obtained from the real game as it is done in MuZero. Then, we traverse the tree built when playing and pick the action with the highest visit count for each node. This would give us the action history \(a_t^0, ..., a_{t}^{N}\) where \(N\) is the length of this path. As the planning is done with hidden states and not with the environment, we do not have direct access to the real rewards of the simulated path. Besides that, there are no guarantees that the greedy simulated path would be the same as the trajectory from the real game. Therefore, we cannot use the rewards from the real game for the simulated path. After obtaining the action history, we query the environments for the rewards by applying in order the action history to the environment, thus obtaining the reward history  \(u_t^0, ..., u_{t}^N \).  After obtaining the rewards above, the value history \(z_t^0, ..., z_{t}^N\)  can then be calculated. First, we obtain the values \(v^0_t,...,v^N_t\) for the nodes along the path and then use those values and the reward history to calculate the n-step return.
Algorithm \ref{alg:traj} shows how to integrate the steps above during training.

We are now querying the environment to obtain the rewards. The impact on the performance is not significant as it is only done for one short path and not for the whole tree. However, this assumes that the environment is reversible, which in many real-world cases is not. Our approach can be considered a middle ground between AlphaZero and MuZero since we query the environment, but way less than in AlphaZero.

Besides that, since AlphaZero and MuZero use a replay buffer, one can consider them off-policy. However, we adopt the same criterion as Willemsen and co-authors \cite{Willemsen2020} and consider MuZero and AlphaZero on-policy since the behaviour and training policy are the same.
\\

\begin{figure}[h]
\centering
\includegraphics[width=0.3\textwidth]{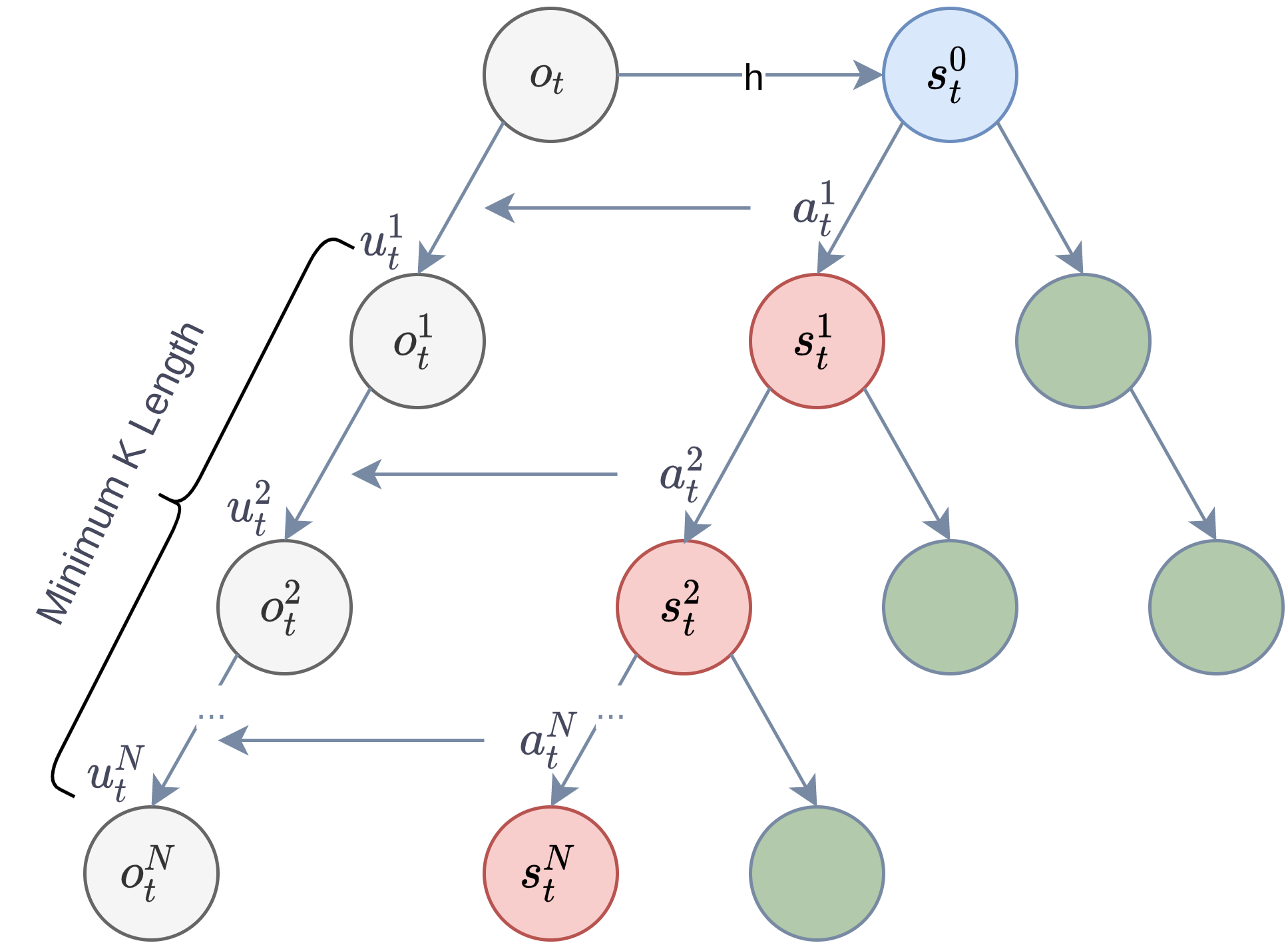}
\caption{How we obtain the values needed to train when using simulated games. We first pick the best actions according to the visit count (the red path). Then, we unroll the environment by applying these actions to the environment obtaining the rewards needed to train.}
\label{fig:off_value_obtaining}
\end{figure}

\noindent
\textbf{Greedy path length.} Let us consider a greedy path as shown in Figure \ref{fig:off_value_obtaining} that we obtained by performing the steps mentioned above. This path will have a length \(N\) meaning that it starts at \(s^0\) and ends at \(s^N\). There are two things we should be aware of regarding the length of this greedy path. 

First, there is no guarantee that its length \(N\) will be the same as the number of unroll steps \(K\). We guarantee that the greedy path has a minimum length of \(K\) by performing more simulations on the leaf node \(s^N\) if needed. 

Secondly, for a state \(s^t\) the value target \(z^t\) is calculated by the n-step return \(z^{t} = u^{t+1} + \gamma u^{t+2} + .. +  \gamma^{n-1}  u^{t+n} + \gamma^{n}v^{t+n}\), where \(n\) is the look-ahead length. We guarantee that the paths have a minimum of \(K\) length. However \(n\) can be larger than \(K\) and performing more simulations on the leaf node to guarantee a minimum length of \(n\) can be too costly. Therefore, we use a variable look-ahead length when training. For example, for a state \(s^t\), if \(t\) plus the look-ahead length is larger than the path length \(N\), we use \(t - N\) as the look-ahead length.
Thus, as we calculate the value targets along the path, the look-ahead length will decrease.

If we only use data from simulated games, we have an algorithm that meets all the conditions of the deadly triad \cite{Sutton2018}: it has function approximation, it bootstraps and it is off-policy. Thus, there is no guarantee of convergence. Furthermore, as we go along a greedy path from a simulated game, the number of visits of each node decreases, and therefore the quality of the targets decreases. This, combined with the decrease of the look-ahead length when calculating the value target, might affect the ability to converge. We address this by combining trajectories from real games with the trajectories from simulated games.

\begin{algorithm}[ht]
\footnotesize
\DontPrintSemicolon
game = newGame()

gameHistory = gameHistory()

done = False
\BlankLine
\While{not done}{
    root = MCTS.simulate(game.observations())

    action = MCTS.selectAction(root.childVisits)

    observation, reward, done = game.step(action)
    \BlankLine
    //traverse the tree greedily and do more simulations if needed
    
    greedyActions, leaf = MCTS.traverseGreedily(root, game.observations())
    \BlankLine
    gameCopy = game.clone()
    
    offPolicyTrajectory = []
    
    //Create off policy trajectory by applying the greedy actions to the environment
    \BlankLine
    \For{$j\gets0$ \KwTo numberOfUnrollSteps}{
        observation, reward, \_ = gameCopy.step(action[$j$])
        
        offPolicyTrajectory.append((observation, reward, action[$j$]))
    } 
    
    gameHistory.save(root, observation, reward, action, done, offPolicyTrajectory)
}
\BlankLine
addTrajectoriesToBuffer(replayBuffer, gameHistory)

\caption{How the data for a trajectory is obtained}
\label{alg:traj}
\end{algorithm}

\subsection{Combining Off and On-policy Targets} \label{ssec:decouple}
The loss for an observation as defined in Equation \ref{eq:mz_loss} can be separated into two distinct components: the value component which includes the reward and value loss; and the policy component which includes the policy loss. 
\begin{equation}
\footnotesize
l_t(\theta) = l_{value} + l_{policy},
\end{equation}
\begin{equation}
\footnotesize
 l_{value} =  \sum^{k}_{k=0} \Big[ l^r(u_{t+k}, r^{k}_{t}) + l^v(z_{t+k}, v^k_{t}) \Big],
\end{equation}
\begin{equation}
\footnotesize
l_{policy} = \sum^{k}_{k=0}  l^p(\pi_{t+k}, p^k_{t}).
\end{equation}

Note that we joined the reward loss and value loss into one component since their values are intertwined because we calculate the value based on the rewards.

Let \(l_{t}^{real}\) be the loss for real game trajectory, and \(l_{t}^{simulated}\) the loss for the greedy trajectory from simulated games.
We can combine these two trajectories and their targets as follows:

\begin{equation}
\footnotesize
l^{combined}_t(\theta) = \alpha l^{real}_{value} + \beta l^{real}_{policy} + \gamma l^{simulated}_{value} + \delta l^{simulated}_{policy},
\end{equation}
where \(\alpha\), \(\beta\), \(\gamma\), \(\delta\) are parameters to control the influence of each respective loss. 

If \(\gamma = \delta = 0\), the loss is the same as the one used in MuZero, while if \(\alpha = \beta = 0\), we only use trajectories from simulated games. This would make MuZero an off-policy algorithm.

If \(\alpha = \delta = 0\), we have something similar to A0GB where we use the greedy value and the policy from real games.
\\

\noindent 
\textbf{Scaling.} \label{scaling} We would like the loss to have the same magnitude regardless of the parameters used. We do this by normalizing the parameters used in the value and policy losses, using Equation \ref{eq:scale_val} and Equation \ref{eq:scale_pol} respectively.
\begin{align}
\footnotesize
    \alpha^{'} = \frac{\alpha}{\alpha + \gamma} && \gamma^{'} = \frac{\gamma}{\alpha + \gamma} 
    \label{eq:scale_val}
\end{align}
\begin{align}
\footnotesize
    \beta^{'} = \frac{\beta}{\beta + \delta} && \delta^{'} = \frac{\delta}{\beta + \delta}
    \label{eq:scale_pol}
\end{align}
\section{Experimental Evaluation}
\label{sec:eval}
The proposed extension is built over an open-source implementation of MuZero \cite{muzerocode}. We are mainly concerned with two things: the convergence speed and how high the final reward is.

There are two types of environments that we can consider for MuZero: two-player zero-sum games and games that behave like general MDPs. Due to the computational resources needed to run MuZero, the environments chosen needed to be relatively simple. 
We decided to use three environments: Cartpole, which has intermediate rewards; MiniGrid, which has sparse rewards; and TicTacToe, a two-player zero-sum game.

\noindent

For all environments, we tested several parameter combinations which are as follows: M0OFF only uses the off-policy targets; M0GB uses the off-policy value target and on-policy policy target, similarly to A0GB; M0OFFV uses all on-policy targets and the off-policy value target; M0ALL uses all possible targets. The values for these combinations are presented in Table \ref{tab:results}.

In this and other tables,  the values of the parameters are shown before the scaling described in Equations \ref{eq:scale_val} and \ref{eq:scale_pol}.

In all plots, the shaded area represents the minimum and maximum values observed and the line is the mean.  We also used smoothing for these plots.

\begin{table*}[t]
  \centering
\footnotesize

\begin{tabular}{|l|llll|l|l|l|l|l|l|l|l|}
\hline
\multirow{2}{*}{} & \multicolumn{4}{l|}{\begin{tabular}[c]{@{}l@{}}Parameters \\ Combinations\end{tabular}} & \multicolumn{2}{l|}{Cartpole}                                               & \multicolumn{2}{l|}{TicTacToe}                                      & \multicolumn{4}{c|}{MiniGrid}                         \\ \cline{2-13} 
                  & \(\alpha\)           & \(\beta\)           & \(\gamma\)           & \(\delta\)          & Reward      & \begin{tabular}[c]{@{}l@{}}Steps \\ until solved\end{tabular} & Reward & \begin{tabular}[c]{@{}l@{}}Opponent \\ Reward\end{tabular} & 3x3         & 4x4         & 5x5         & 6x6         \\ \hline
MuZero            & 1                    & 1                   & 0                    & 0                   & \textbf{306} \(\pm\) 136 & 4148                                                 &6.51 \(\pm\) 4.95     & 6.20 \(\pm\)4.66                         & 9.80 \(\pm\) 0.18 & 9.48 \(\pm\) 0.61 & 9.10 \(\pm\) 0.80 & 8.73 \(\pm\) 1.20 \\ \hline
M0OFF             & 0                    & 0                   & 1                    & 1                   & 123 \(\pm\) 74 & Not solved                                                    & 1.22 \(\pm\) 2.31      & 14.50 \(\pm\) 4.34                      & 1.51 \(\pm\) 1.80 & 1.23 \(\pm\) 0.84 & 0.75 \(\pm\) 0.73 & 0.26 \(\pm\) 0.52 \\ \hline
M0GB              & 0                    & 1                   & 1                    & 0                   & 178 \(\pm\) 92 & Not solved                                                     & 2.54 \(\pm\) 2.80      & 10.55 \(\pm\) 4.81                      & 9.67 \(\pm\) 0.44 & 5.81 \(\pm\) 4.8  & 1.36 \(\pm\) 0.67 & 0.44 \(\pm\) 0.76 \\ \hline
M0OFFV            & 1                    & 1                   & 1                    & 0                   & 250 \(\pm\) 131 & 3745                                                          & 5.44 \(\pm\) 4.52      & 6.03 \(\pm\) 5.10                      & \textbf{9.81} \(\pm\) 0.17 & 9.52 \(\pm\) 1.25 & 9.20 \(\pm\) 1.20 & 8.88 \(\pm\) 1.33 \\ \hline
M0ALL             & 1                    & 1                   & 1                    & 1                   & 223 \(\pm\) 93 & \textbf{3348}                                                  & 3.89 \(\pm\) 3.96      & 10.21 \(\pm\) 5.25                      & 9.45 \(\pm\) 0.70 & 7.67 \(\pm\) 3.02 & 7.78 \(\pm\) 2.67 & 7.40 \(\pm\) 2.34 \\ \hline
Decaying Value   & 1                    & 1                   & -             & 0                   & 253 \(\pm\) 121 & 6648                                                    & \textbf{7.23} \(\pm\) 5.49      & \textbf{4.85} \(\pm\) 4.38                  & 9.74 \(\pm\) 0.59 & \textbf{9.64} \(\pm\) 0.45 & \textbf{9.39} \(\pm\) 0.83 & \textbf{9.18} \(\pm\) 0.97 \\ \hline
Decaying Value and Policy                  & 1                    & 1                   & -             & -            & 295 \(\pm\) 143 & 5848                                                  & -      & -                         & - & - & - & - \\ \hline

\end{tabular}
  \caption{Results for the several games tested.}
  \label{tab:results}
\end{table*}

\subsection{Cartpole} 
This environment consists of a pole attached to a cart placed on a track. The objective is to keep the pole that starts upright from falling over. A reward of +1 is given for every timestep that the pole remains upright. The environment is considered solved if an average reward of 195 is obtained over 100 episodes. Episodes end when the pole is more than 15 degrees from the vertical axis, or the cart moves more than 2.4 units from the center of the track.

At each timestep, there are only two possible actions: to move right or to move left. An observation consists of four components: the cart position, the cart velocity, the pole angle, and the pole angular velocity.

\textbf{Parameters.}
We did 10 runs for each parameter combination. We used an n-step of 50, an unroll step of 10, and a simulation number of 50. The hidden state is of size 8, the observations of size 4, and the action size is 2. We used a support of size 21. We trained for 25000 steps with a batch size of 128 and a replay buffer size of 500 games. A training step consists of training the network with one batch. 

Besides the episode termination conditions presented above, we also guarantee that episodes do not have more than 500 steps. Therefore, the maximum reward of an episode is 500. 

\begin{figure}
\centering
	\includegraphics[width=0.5\textwidth]{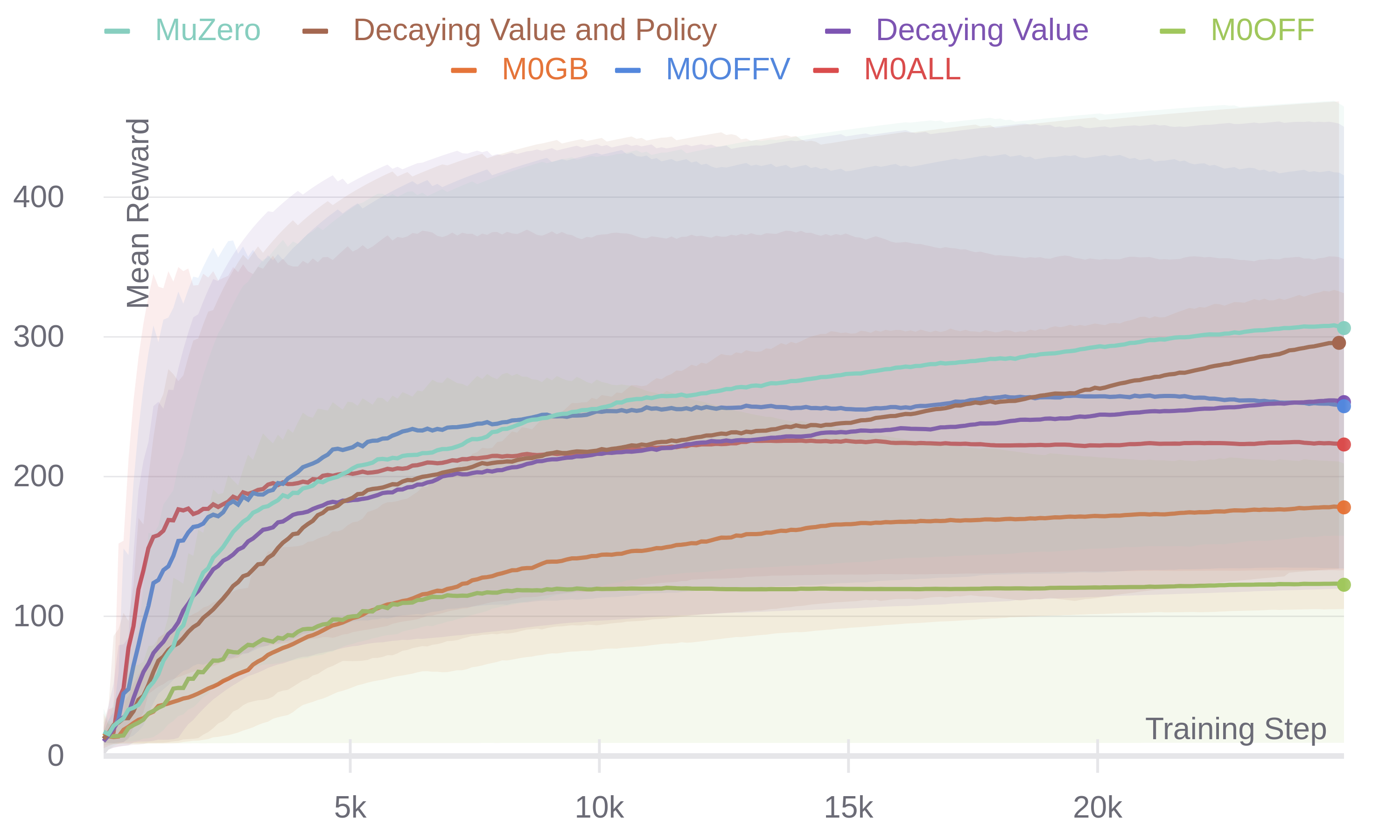}
	\caption{Cartpole results for the several parameters tested. }
	\label{fig:cartpole_main} 	
\end{figure}

\textbf{Results.}
In Table \ref{tab:results} we can see that, for the parameters tested, MuZero is able to achieve the highest end reward of 306, followed by M0OFFV with a reward of 250. The lowest reward comes from training completely with simulated games with a reward of 123.
Figure \ref{fig:cartpole_main} and Table \ref{tab:results} show that M0ALL solves the environment the quickest in 3348 steps, followed by M0OFFV in 3745. However, both these parameter combinations only achieve end rewards of 223 and 250, respectively. Besides that, they also converge quickly to these end reward values. M0OFFV reaches values close to its end reward at around 8400 steps, and M0ALL at around 6000 steps. 
MuZero continues learning after solving the environment at training steps 4148. M0OFF and M0GB are not able to solve the environment. 


Based on the results above, we decided to train using the off-policy targets and then decay the values for those targets. We did runs with both \(\gamma\) and \(\delta\) decaying, and also, runs with only \(\gamma\) decaying and \(\delta\) set to zero. In this way, we can take advantage of the fast initial convergence speeds that characterize runs that use the off-policy targets without having training stagnate towards the end.

We trained for 25000 steps using the parameter decay in Table \ref{tab:decay}. Figure \ref{fig:cartpole_main} shows that runs with decaying parameters do not seem to provide any clear benefit compared to MuZero. Runs with decaying value and policy achieved similar values to MuZero with an end reward of 295 and solve the environment at around 5848 steps. While runs with decaying value achieved an end reward of 253 and solve the environment at around 6648 training steps. 


\subsection{TicTacToe}
We used a board of size 3x3. When a game is over, the winner gets a reward of 20, and the loser a reward of -20. If the game is a draw, the reward is zero.

\textbf{Parameters.}
We did 10 runs for each parameter combination. We used an n-step of 9, an unroll step of 3, and a simulation number of 25. The hidden state is of size 8, the observations of size 9, and the action size is 9. We used a support of size 1. We trained for 25000 steps with a batch size of 64 and a replay buffer size of 3000 games.

\begin{figure}
 	\includegraphics[width=0.5\textwidth]{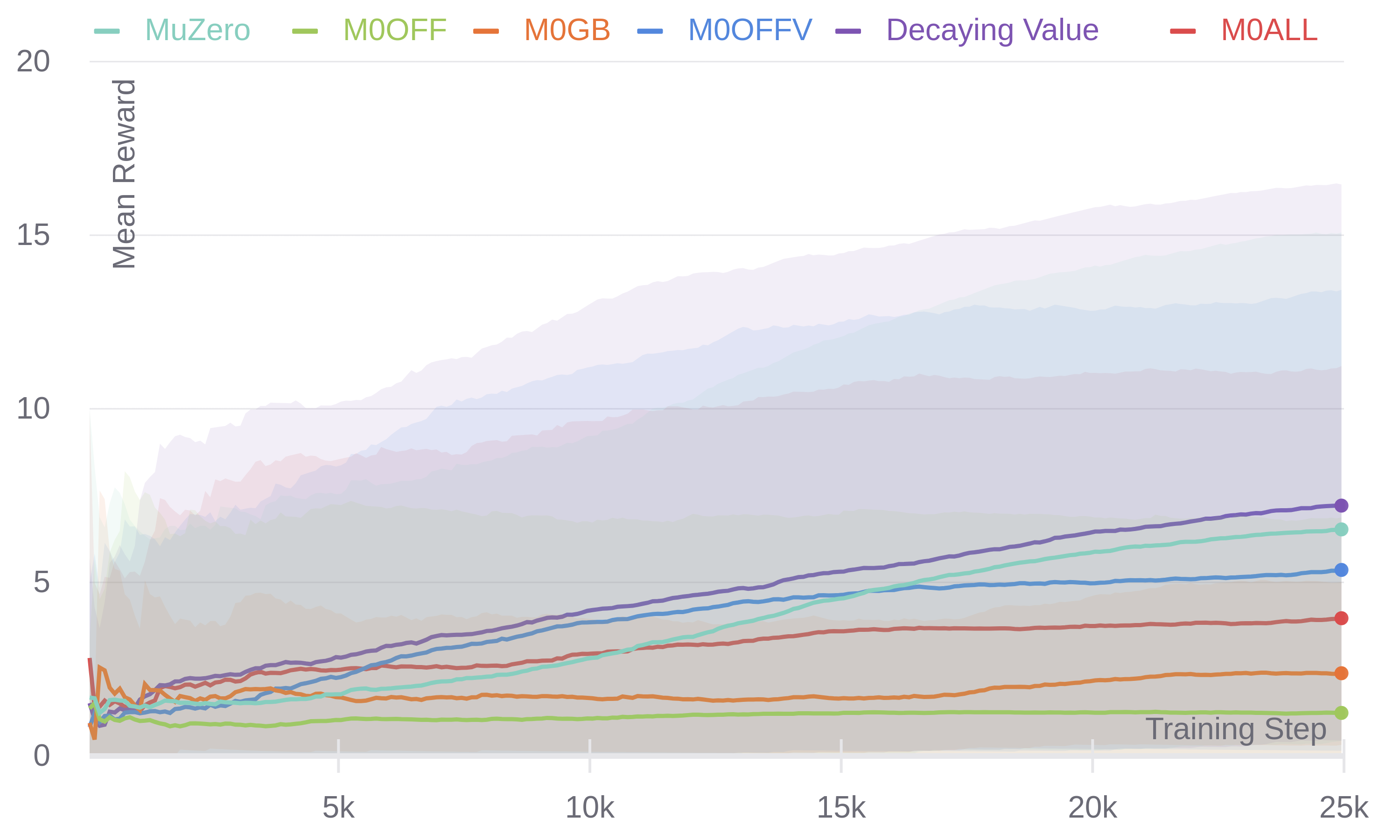}
	\caption{TicTacToe Results.}
	\label{fig:tic_results}
\end{figure}

\textbf{Results.}
In Table \ref{tab:results}, MuZero achieves the highest reward with a reward of 6.51, followed by M0OFFV with a reward of 5.44. In Figure \ref{fig:tic_results}, we can see that M0OFFV converges faster than MuZero initially but then learning slows down, converging to a lower reward. Both M0GB and M0OFF are not able to converge. Contrary to Cartpole, M0ALL performed worse than M0OFFV.

We trained with decaying parameters, using only the off-policy value target since M0ALL performed worse than M0OFFV. We decay the parameters according to Table \ref{tab:decay}. As we can see in Figure \ref{fig:tic_results}, using decaying parameters, we were able to achieve faster convergence and higher rewards than MuZero, with an end reward of 7.23 and opponent reward of 4.85.

\subsection{MiniGrid}
This environment consists of an \(N \times N\) grid. The agent starts on the top left corner, and the objective is to reach the opposite corner. A reward of +10 is given when the agent reaches the corner. Episodes end when the agent reaches the corner or when \(N + N\) steps have passed. Therefore, the agent has to play optimally, or it does not get any reward at all, since the shortest path to the opposite corner is \(N + N\). 

At each timestep, there are only two possible actions: to move down or to move right. An observation consists of the whole grid \(N \times N \).

\textbf{Parameters.}
We used n-step of 7, unroll step of 7, and a simulation number of 5. The hidden state is of size 5, the observations of size \(N \times N\), and the action size of size 2. We used a support of size 21, a batch size of 32 and a replay buffer size of 5000 games. In this environment, the grid has size \(N \times N\).  

We ran experiments with \(N = {3,4,5,6}\) and did 6 runs for each parameter combination. For \(N = 3,4\) we ran for 15000 training steps and for \(N = 5,6\) we ran for 20000 training steps. The other parameters are the same for all \(N\).

\textbf{Results.}
Table \ref{tab:results} shows that as we increase \(N\), the gap between the model with the best result and others increases. For all \(N\), M0OFFV is able to achieve higher end rewards than MuZero.  

\begin{figure}
    \centering
	\includegraphics[width=0.5\textwidth]{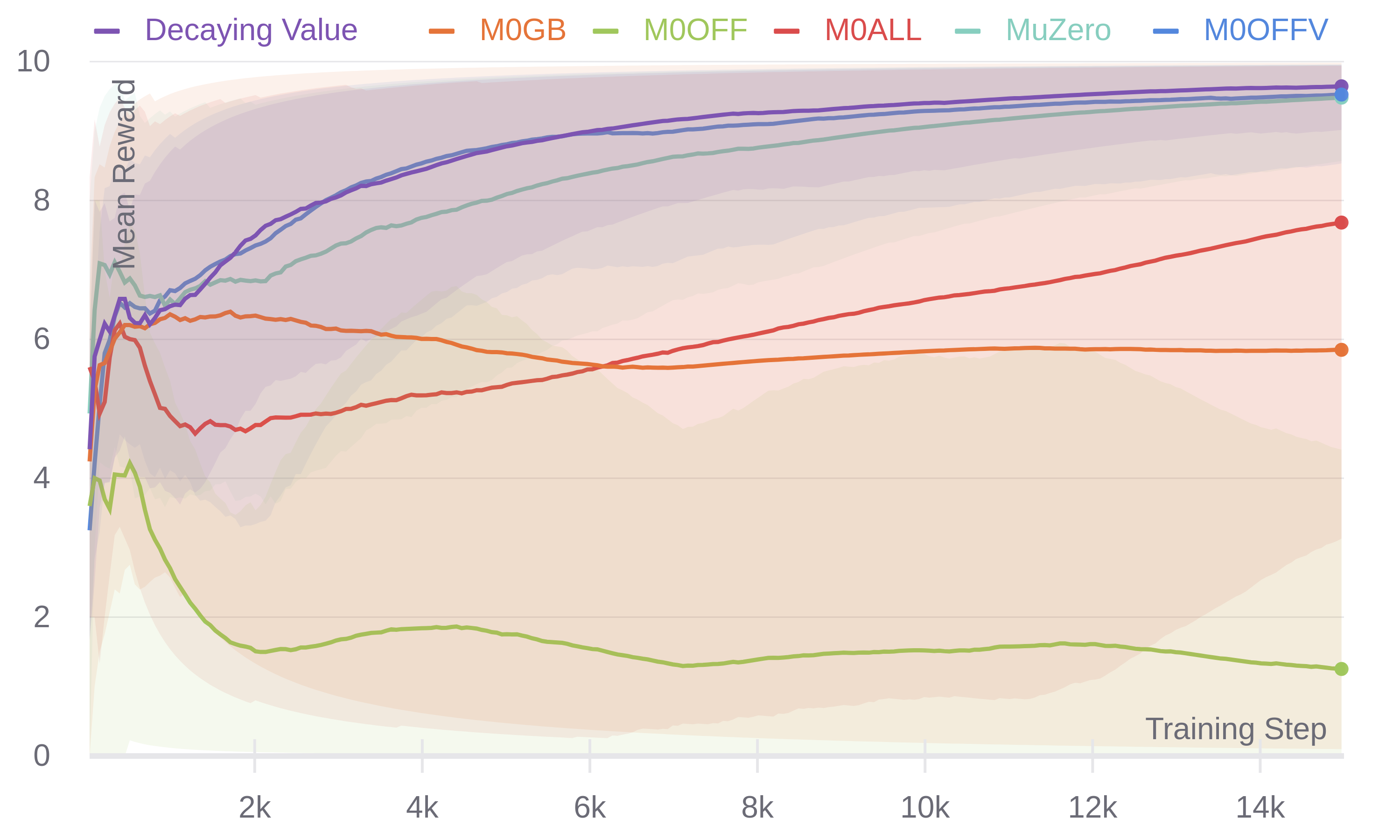}
	\caption{4x4 MiniGrid Results.}
	\label{fig:4x4res}
\end{figure}

\begin{figure}
    \centering
	\includegraphics[width=0.5\textwidth]{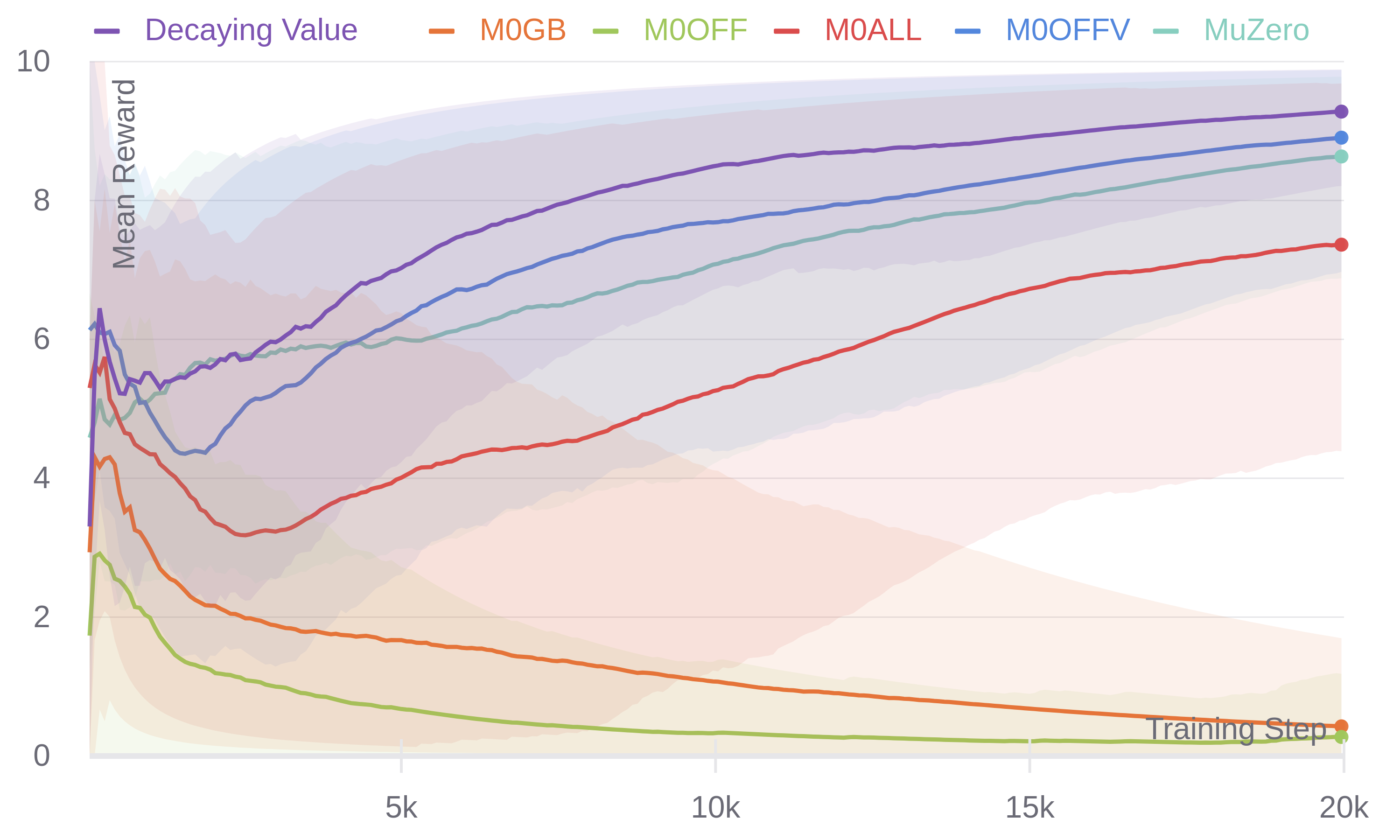}
	\caption{6x6 MiniGrid Results.}
	\label{fig:6x6res}
\end{figure}

Figures \ref{fig:4x4res} and \ref{fig:6x6res} show how the training progresses for grid size \(N = 4\) and \(N = 6\), respectively. Besides reaching higher final rewards, M0OFFV seems to converge faster than MuZero consistently.
M0GB was only able to converge to a high final reward, of 9.67, for \(N = 3\). For \(N = 4\), it obtained a mean final reward of 5.81 with a variance of 4.8, meaning that it sometimes was able to converge to high rewards and other time it didn't converge. For \(N = 5\) and \(N = 6\) it was not able to converge.
M0OFF does not converge when using only data from simulated games, a behaviour already observed in Cartpole and TicTacToe. 

We trained with decaying parameters, and similarly to TicTacToe, since M0ALL performed worse than M0OFFV, we only used the off-policy value target. We decay the parameters according to Table \ref{tab:decay}. As we can see in Figure \ref{fig:4x4res} and \ref{fig:6x6res}, runs that use decaying parameters converged faster and were able to reach higher end rewards than MuZero and M0FFV. 


\begin{table*}[t]
  \centering
\footnotesize
\begin{tabular}{|llll|llll|l|llll|l|llll|l|}
\hline
\multicolumn{9}{|l|}{Catpole}                                                                                        & \multicolumn{5}{l|}{\multirow{2}{*}{TicTacToe}}                   & \multicolumn{5}{l|}{\multirow{2}{*}{MiniGrid}}                    \\ \cline{1-9}
\multicolumn{4}{|l|}{Decaying Value}             & \multicolumn{5}{l|}{Decaying Value and Policy}                    & \multicolumn{5}{l|}{}                                             & \multicolumn{5}{l|}{}                                             \\ \hline
\(\alpha\) & \(\beta\) & \(\gamma\) & \(\delta\) & \(\alpha\) & \(\beta\) & \(\gamma\) & \(\delta\) & Training Steps & \(\alpha\) & \(\beta\) & \(\gamma\) & \(\delta\) & Training Steps & \(\alpha\) & \(\beta\) & \(\gamma\) & \(\delta\) & Training Steps \\ \hline
1          & 1         & 1          & 0          & 1          & 1         & 1          & 1          & 0 - 6249       & 1          & 1         & 1          & 0          & 0 - 6249       & 1          & 1         & 1          & 0          & 0 - 4999       \\ \hline
1          & 1         & 0.5        & 0          & 1          & 1         & 0.5        & 0.5        & 6250 - 12499   & 1          & 1         & 0.5        & 0          & 6250 - 12499   & 1          & 1         & 0.5        & 0          & 5000 - 9999    \\ \hline
1          & 1         & 0          & 0          & 1          & 1         & 0          & 0          & 12500+         & 1          & 1         & 0          & 0          & 12500+         & 1          & 1         & 0          & 0          & 10000+         \\ \hline
\end{tabular}
  \caption{How the parameters change along training for the several games tested.}
  \label{tab:decay}
\end{table*}



\subsection{Analysis}
\label{sec:anal}
\textbf{Off-policy value target \(\gamma\).}
Across all environments, the usage of this target provides faster initial convergence speeds. However, if we compare M0FFV runs with M0GB runs, we can see that using solely this target for the value is not enough, and it needs to be combined with the on-policy value target. 

In Cartpole and TicTacToe, despite faster initial convergence speeds, we can see that M0FFV runs stagnate towards the end, reaching a lower end reward than MuZero. 
In MiniGrid, however, this value target was enough to provide faster initial convergence speeds and to achieve higher end rewards.

In environments with sparse rewards, runs that used this target with decaying parameters were able to have a faster initial convergence with rewards higher than MuZero.

\textbf{Off-policy policy target \(\delta\).}
This target only provided benefits in Cartpole. In this environment, runs that used this target have fast initial convergence, but then quickly stagnate to values lower than MuZero, which leads us to posit that this target does not seem to improve training and impairs convergence. This target is more dependent on the number of simulations than the value target because it is calculated solely based on the visit count, and as we go along a simulated trajectory, the visit count decreases. However, based on the M0ALL training curve, this target seems to be useful in the beginning, as M0OFFV does not have it and shows slower convergence speeds. 

\begin{figure}
	\includegraphics[width=0.5\textwidth]{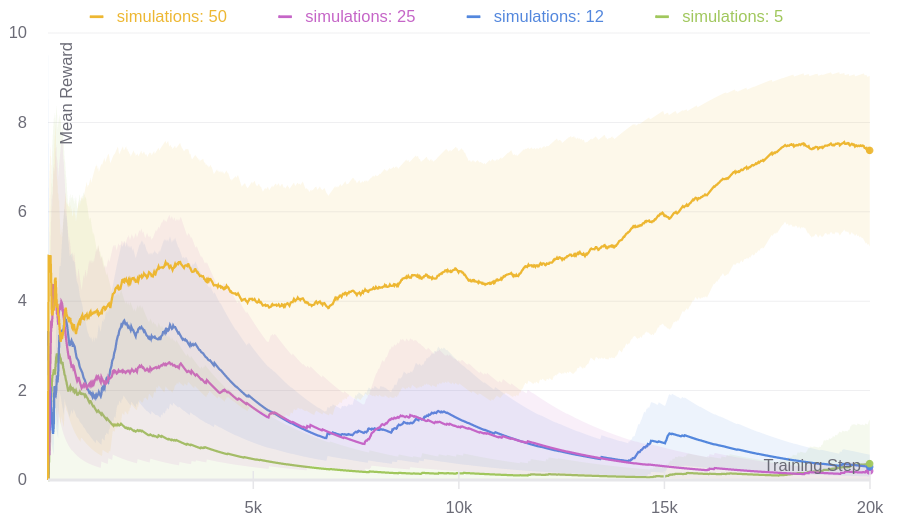}
	\caption{Mean reward for M0OFF runs with varying simulation size for 6x6 MiniGrid.}
	\label{fig:off_sim}
\end{figure}

In MiniGrid and TicTacToe, this target does not seem to provide any benefit, most likely due to the lower MCTS simulation count used. Figure \ref{fig:off_sim} shows how off-policy targets are dependent on the number of simulations in MiniGrid. We can see that M0OFF needed 50 simulations to be able to converge. The final mean reward obtained by M0OFF with this number of simulations is still lower than the one obtained with other targets, shown in Table \ref{tab:results}, which only uses 5 simulations.

\textbf{Comparing environments.}
These new targets did not perform as well on Cartpole as they did on TicTacToe or MiniGrid.

For the on-policy value target, we posit that this might be due to the fact that these are environments with sparse rewards, where there are no intermediate rewards generated by the environment that can be used to help training. Therefore, the usage of the off-policy value target provides a clear benefit. On the other hand, in an environment with intermediate rewards, like Cartpole, the on-policy value target provides a higher quality training signal due to the fact that its value is calculated using intermediate rewards that come from the environment. Therefore, the off-policy value target might not provide any new information. 

Besides that, in Figure \ref{fig:cartpole_main} we can see that the M0ALL and M0OFFV have faster initial convergence that, however, quickly stagnate in values lower than MuZero. We think that these off-policy targets, in Cartpole, are leading the model to find and quickly overfit to a suboptimal strategy. As in the runs with decaying value and policy, we can see that the result also stagnates, but towards the end, when \(\gamma\) and \(\delta\) approach zero, the model starts to learn again and reaches values closer to MuZero.
We conjecture that a better parameter scheduling strategy might make these targets more useful in this environment.

\section{Conclusions and Future Work}
\label{sec:con}
In this work, we presented a way to obtain off-policy targets by using data from MCTS simulations. We then combined these targets with on-policy targets in multiple ways and performed experiments in three environments with distinct characteristics. 

The results show that off-policy targets can be useful for training and can speed up convergence. 

In environments with sparse rewards, runs that use the off-policy value target with decay are able to have faster initial convergence and achieve higher rewards than MuZero. 

In the case of Cartpole, an environment with intermediate rewards, these runs were able to provide a faster initial convergence than MuZero, but stagnated to lower values. We think that these runs are overfitting a suboptimal strategy and conjecture that using a better parameter scheduling strategy would make the targets more beneficial during training. 

The usefulness of the off-policy policy target seems to be dependent on the number of simulations used.

One thing to also note, is that if we only consider a version of the changes presented in this work that is only concerned with environments with sparse rewards, we no longer need to query the environment to obtain the intermediate rewards. Therefore, we have an algorithm that converges faster than MuZero and no longer needs to assume that the environment is reversible.

\subsection{Future Work}
There were some trade-offs we made due to the variable size of the simulated trajectories, detailed in \ref{sec:creating_sim}. 
\begin{itemize}
    \item When the path length is smaller than the number of unroll steps, we choose to perform simulations on the leaf node of that path. However, we could perform simulations on the root node, making us discover another better trajectory. We could also divide our simulation budget as such: if we have a budget of 30 simulations and need more 3 nodes to reach the unroll step size, we can perform 10 simulations on the leaf node, then 10 on the next leaf node, etc. This would make us have higher quality samples for the leaf nodes. The approach we choose is the middle ground between these two. However, a study of these approaches would be useful.
    \item We choose to use a variable look-ahead length. However, we could use a small simulation budget to make the path reach the look-ahead length size.
\end{itemize}

GradNorm \cite{Chen2018} balances the losses of networks that are optimizing for several tasks in real-time based on the gradients of those losses. The results show that the best parameters depend on the type of environment and that using parameters that change during training seems to provide the best results. Therefore, using GradNorm to compute \(\alpha, \beta, \delta, \gamma\) during training would be useful since we would not have to find the best values for these parameters for each environment.

Recent work \cite{Grill2020} has identified several problems that exist due to the fact that we use the visit count to calculate the policy target MCTS \(\pi_{MCTS}\) (Equation \ref{eq:pickmove}).
\begin{itemize}
    \item Visit counts take too long to reflect newly discovered information. For example, if we get an unexpected win or loss, we need to do more simulations before this is reflected in the visit counts.
    \item Low visit counts can only express a limited subset of policies. For example, if we only do 5 simulations, then all policy targets are necessarily a multiple of 1/5.
    \item Using visit count as a policy target does not mean much for actions that were not visited once. We have gained no new information about that state, and now we will try to lower its policy value even further.
\end{itemize}
The policy target could be calculated using the regularized Q-values of each child node \cite{Grill2020}. This new policy target performed better than using visit counts, especially for environments with a low number of simulations. Therefore, calculating the policy target from the regularized Q-values instead of the visit counts of each child node could significantly improve the usefulness of the off-policy policy target since the usefulness of this target seems to be correlated with the number of simulations used.

\label{chap:conclusion}

\bibliographystyle{IEEEtran}

{\footnotesize\bibliography{references}}

\begin{thebibliography}{10}
\providecommand{\url}[1]{#1}
\csname url@samestyle\endcsname
\providecommand{\newblock}{\relax}
\providecommand{\bibinfo}[2]{#2}
\providecommand{\BIBentrySTDinterwordspacing}{\spaceskip=0pt\relax}
\providecommand{\BIBentryALTinterwordstretchfactor}{4}
\providecommand{\BIBentryALTinterwordspacing}{\spaceskip=\fontdimen2\font plus
\BIBentryALTinterwordstretchfactor\fontdimen3\font minus
  \fontdimen4\font\relax}
\providecommand{\BIBforeignlanguage}[2]{{%
\expandafter\ifx\csname l@#1\endcsname\relax
\typeout{** WARNING: IEEEtran.bst: No hyphenation pattern has been}%
\typeout{** loaded for the language `#1'. Using the pattern for}%
\typeout{** the default language instead.}%
\else
\language=\csname l@#1\endcsname
\fi
#2}}
\providecommand{\BIBdecl}{\relax}
\BIBdecl

\bibitem{Silver2016}
D.~Silver, A.~Huang, C.~J. Maddison, A.~Guez, L.~Sifre, G.~van~den Driessche,
  J.~Schrittwieser, I.~Antonoglou, V.~Panneershelvam, M.~Lanctot, S.~Dieleman,
  D.~Grewe, J.~Nham, N.~Kalchbrenner, I.~Sutskever, T.~Lillicrap, M.~Leach,
  K.~Kavukcuoglu, T.~Graepel, and D.~Hassabis, ``Mastering the game of go with
  deep neural networks and tree search,'' \emph{Nature}, vol. 529, 12 2016.

\bibitem{Silver2017}
D.~Silver, H.~Hasselt, M.~Hessel, T.~Schaul, A.~Guez, T.~Harley,
  G.~Dulac-Arnold, D.~Reichert, N.~Rabinowitz, A.~Barreto \emph{et~al.}, ``The
  predictron: End-to-end learning and planning,'' \emph{International
  Conference on Machine Learning}, pp. 3191--3199, 2017.

\bibitem{Silver2018}
D.~Silver, T.~Hubert, J.~Schrittwieser, I.~Antonoglou, M.~Lai, A.~Guez,
  M.~Lanctot, L.~Sifre, D.~Kumaran, T.~Graepel, T.~Lillicrap, K.~Simonyan, and
  D.~Hassabis, ``A general reinforcement learning algorithm that masters chess,
  shogi, and go through self-play,'' \emph{Science}, vol. 362, 12 2018.

\bibitem{Schrittwieser2020}
J.~Schrittwieser, I.~Antonoglou, T.~Hubert, K.~Simonyan, L.~Sifre, S.~Schmitt,
  A.~Guez, E.~Lockhart, D.~Hassabis, T.~Graepel, T.~Lillicrap, and D.~Silver,
  ``Mastering atari, go, chess and shogi by planning with a learned model,''
  \emph{Nature}, vol. 588, 12 2020.

\bibitem{Coulom2007}
R.~Coulom, ``Efficient selectivity and backup operators in monte-carlo tree
  search,'' \emph{Lecture Notes in Computer Science (including subseries
  Lecture Notes in Artificial Intelligence and Lecture Notes in
  Bioinformatics)}, vol. 4630 LNCS, 2007.

\bibitem{Willemsen2020}
D.~Willemsen, H.~Baier, and M.~Kaisers, ``Value targets in off-policy
  alphazero: a new greedy backup,'' \emph{Adaptive and Learning Agents (ALA)
  Workshop}, 2020.

\bibitem{Tesauro1995}
G.~Tesauro, ``Temporal difference learning and td-gammon,''
  \emph{Communications of the ACM}, vol.~38, pp. 58--68, 1995.

\bibitem{Campbell2002}
M.~Campbell, A.~J. Hoane, and F.~H. Hsu, ``Deep blue,'' \emph{Artificial
  Intelligence}, vol. 134, 2002.

\bibitem{Gu2017}
S.~S. Gu, T.~Lillicrap, R.~E. Turner, Z.~Ghahramani, B.~Schölkopf, and
  S.~Levine, ``Interpolated policy gradient: Merging on-policy and off-policy
  gradient estimation for deep reinforcement learning,'' \emph{Advances in
  neural information processing systems}, pp. 3846--3855, 2017.

\bibitem{Gu2016}
S.~Gu, T.~Lillicrap, Z.~Ghahramani, R.~E. Turner, and S.~Levine, ``Q-prop:
  Sample-efficient policy gradient with an off-policy critic,'' \emph{arXiv
  preprint arXiv:1611.02247}, 2016.

\bibitem{Lehnert2015}
L.~Lehnert and D.~Precup, ``Policy gradient methods for off-policy control,''
  \emph{arXiv preprint arXiv:1512.04105}, 2015.

\bibitem{Hu2019}
K.-C. Hu, C.-H. Pi, T.~H. Wei, I.~Wu, S.~Cheng, Y.-W. Dai, W.-Y. Ye
  \emph{et~al.}, ``Towards combining on-off-policy methods for real-world
  applications,'' \emph{arXiv preprint arXiv:1904.10642}, 2019.

\bibitem{Wang2013}
Y.-H. Wang, T.-H.~S. Li, and C.-J. Lin, ``Backward q-learning: The combination
  of sarsa algorithm and q-learning,'' \emph{Engineering Applications of
  Artificial Intelligence}, vol.~26, pp. 2184--2193, 2013.

\bibitem{Rummery1994}
G.~A. Rummery and M.~Niranjan, \emph{On-line Q-learning using connectionist
  systems}.\hskip 1em plus 0.5em minus 0.4em\relax University of Cambridge,
  Department of Engineering Cambridge, UK, 1994, vol.~37.

\bibitem{Watkins1992}
C.~J. C.~H. Watkins and P.~Dayan, ``Q-learning,'' \emph{Machine learning},
  vol.~8, pp. 279--292, 1992.

\bibitem{Hausknecht2016}
M.~Hausknecht and P.~Stone, ``On-policy vs. off-policy updates for deep
  reinforcement learning,'' \emph{Deep Reinforcement Learning: Frontiers and
  Challenges, IJCAI 2016 Workshop}, 2016.

\bibitem{Mnih2013}
V.~Mnih, K.~Kavukcuoglu, D.~Silver, A.~Graves, I.~Antonoglou, D.~Wierstra, and
  M.~Riedmiller, ``Playing atari with deep reinforcement learning,''
  \emph{arXiv preprint arXiv:1312.5602}, 2013.

\bibitem{Lillicrap2015}
T.~P. Lillicrap, J.~J. Hunt, A.~Pritzel, N.~Heess, T.~Erez, Y.~Tassa,
  D.~Silver, and D.~Wierstra, ``Continuous control with deep reinforcement
  learning,'' \emph{arXiv preprint arXiv:1509.02971}, 2015.

\bibitem{Auger2013}
D.~Auger, A.~Couetoux, and O.~Teytaud, ``Continuous upper confidence trees with
  polynomial exploration–consistency,'' \emph{Joint European Conference on
  Machine Learning and Knowledge Discovery in Databases}, pp. 194--209, 2013.

\bibitem{Carlsson2019}
F.~Carlsson and J.~{\"O}hman, ``Alphazero to alpha hero: A pre-study on
  additional tree sampling within self-play reinforcement learning,'' 2019.

\bibitem{Moerland2018}
T.~M. Moerland, J.~Broekens, A.~Plaat, and C.~M. Jonker, ``A0c: Alpha zero in
  continuous action space,'' \emph{arXiv preprint arXiv:1805.09613}, 2018.

\bibitem{Sutton2018}
R.~S. Sutton and A.~G. Barto, \emph{Reinforcement learning: An
  introduction}.\hskip 1em plus 0.5em minus 0.4em\relax MIT press, 2018.

\bibitem{muzerocode}
``Open-source muzero implementation,''
  \url{https://github.com/werner-duvaud/muzero-general}, accessed: 2020-12-31.

\bibitem{Chen2018}
Z.~Chen, V.~Badrinarayanan, C.~Y. Lee, and A.~Rabinovich, ``Gradnorm: Gradient
  normalization for adaptive loss balancing in deep multitask networks,''
  \emph{35th International Conference on Machine Learning, ICML 2018}, vol.~2,
  2018.

\bibitem{Grill2020}
J.-B. Grill, F.~Altché, Y.~Tang, T.~Hubert, M.~Valko, I.~Antonoglou, and
  R.~Munos, ``Monte-carlo tree search as regularized policy optimization,''
  \emph{International Conference on Machine Learning}, pp. 3769--3778, 2020.

\end{thebibliography}

\end{document}